# Natural Language Processing Accurately Categorizes Indications, Findings and Pathology Reports from Multicenter Colonoscopy

Shashank Reddy Vadyala[1], Eric A. Sherer[2]
1. Department of Computational Analysis and Modeling, Louisiana Tech University, Ruston, LA United States
2. Department of Chemical Engineering, Louisiana Tech University, Ruston, LA United States

**Abstract**

**Background:** Colonoscopy is used for colorectal cancer (CRC) screening. Extracting details of the colonoscopy findings from free text in electronic health records (EHRs) can be used to determine patient risk for CRC and colorectal screening strategies.

**Objective:** We developed and evaluated the accuracy of a deep learning model framework to extract information for the clinical decision support system to interpret relevant free-text reports, including indications, pathology, and findings notes.

**Methods:** The Bio-Bi-LSTM-CRF framework was developed using Bidirectional Long Short-term Memory (Bi-LSTM) and Conditional Random Fields (CRF) to extract several clinical features from these free-text reports including indications for the colonoscopy, findings during the colonoscopy, and pathology of resected material. We trained the Bio-Bi-LSTM-CRF and existing Bi-LSTM-CRF models on 80% of 4,000 manually annotated notes from 3,867 patients. These clinical notes were from a group of patients over 40 years of age enrolled in four Veterans Affairs Medical Centers. A total of 10% of the remaining annotated notes were used to train hyperparameter and the remaining 10% were used to evaluate the accuracy of our model Bio-Bi-LSTM-CRF and compare to Bi-LSTM-CRF.

**Results:** Bio-Bi-LSTM-CRF test set accuracy was 93.5% for indications reports, 88.0% for findings reports, 96.5% for pathology reports, 85.0% for number of polyps' entity, 81.0% for size of polyps' entity, 94.0% for locations of polyps entity, and 92.0% for poly removal procedure entity, and 96.5% for colon location of polyps. The accuracy of the Bio-Bi-LSTM-CRF methods between facilities ranged from 92.0% to 94.0%. The Bio-Bi-LSTM-CRF model's overall accuracy was between 8.0% and 13.0% points higher than the Bi-LSTM-CRF method in entity identification. The Bio-Bi-LSTM-CRF model achieved a higher accuracy for all outcomes for all four facilities.

**Conclusion:** Our experiments show that the bidirectional encoder representations by integrating dictionary function vector from Bio-Bi-LSTM-CRF and strategies character sequence embedding approach is an effective way to identify colonoscopy features from EHR-extracted clinical notes. The Bio-Bi-LSTM-CRF model creates new opportunities to identify patients at risk for colon cancer and study their health outcomes.

## Introduction

The amount of complex and varied information (e.g., structured, semi-structured, and unstructured) has grown dramatically during the past century in numerous sectors, including education[1], media[2], power[3], and healthcare[4]. In healthcare, electronic health records (EHRs) contain information in both unstructured and structured formats on patient health, disease status, and care received, which can be useful for epidemiologic and clinical research[5]. In EHRs, unstructured data is documented without standard content specifications, often recorded as free text, and structured data is generally entered into discrete data fields with standardized responses or parameters (e.g., age, weight)[6]. Compared to the structured data within the EHR, free-text (unstructured data) data often conveys more granular and contextual information of clinical events and enhances communication between clinical teams[7]. As such, extracting information from these free-text document resources has considerable potential value in supporting a multitude of aims, from clinical decision support and quality care improvement to the secondary use of clinical data for research, public health, and pharmacovigilance activities[8]. Natural language processing (NLP) can be an efficient way of automatically extracting and arranging this information, making NLP a potentially pivotal technology for enabling quality measurement from EHR data.

## Background

Colonoscopy is a well-established procedure used to detect and prevent colorectal cancer (CRC), the third most common cancer in men and women, which accounts for almost 10% of all cancer-related deaths in the United States[9]. There is evidence that colonoscopy use has risen significantly in recent years, primarily because of increased colonoscopy screening rates [10]. Understanding the dynamics of the changes in colorectal polyps and cancer involving sizes, shapes, and the number of polyps is critical because these are associated with the efficacy of colorectal screening strategies. Manually extracting the detailed findings of a colonoscopy (e.g., number of polyps, polyp sizes, and polyp locations) from a colonoscopy procedure report and connecting with the pathology report (e.g., the histology of polyps found during colonoscopy) is quite time-consuming[14]. Extracting information entity identification from the clinical free text requires scalable techniques such as NLP, machine learning, and neural networks to convert clinical free text into a structured variable for text mining and further information extraction[15-17]. Colonoscopy named entity identification has drawn a considerable amount of research in recent years, and various techniques for identifying entities have been suggested [22]. Conventional methods such as Decision Tree, Support Vector Machines (SVM), Hidden Markov Model (HMM), and Conditional Random Field (CRF) are either rule-based or dictionary-based, which usually need to be manually formulated and for the current datasets only. When the dataset is updated or replaced, updating the rules contributes to high machine costs. This will cause a low recall if the original rules or dictionary are used. The latest advanced techniques are data-driven, including the methods of machine learning and deep learning. In particular, the models of Bi-LSTM-CRF (Bi-Long Short-Term Memory and Conditional Random Field) have been used successfully in the biomedical text to achieve better results and represent some of the methods that are most commonly used[23]. Existing work regarding NLP in connection with CRC and colonoscopy is limited in scope and quantity. Adapting a clinical information retrieval system, Named entity recognition (NER), identified pathology reports consistent with CRC from an electronic medical record system[26]. An application was proposed to automatically identify patients in need of CRC screening by detecting the timing and status of colorectal screening tests mentioned in the electronic clinical narrative documentation [2, 14, 26, 27]. Neither of these studies assesses the colonoscopy procedure's detailed findings, such as the number, shape, and size of polys, due to the problem of entity boundary extraction with different entity lengths and named entity recognition. Understanding the dynamics of the changes in colorectal polyps and cancer, involving sizes, shapes, and the number of polyps, is critical because these are associated with the efficacy of colorectal screening strategies[28].

Named entity recognition (NER), also known as entity identification, entity chunking, and entity extraction, is a process of information extraction that helps to locate and classify named entities mentioned in unstructured and structured text into pre-defined categories such as medical codes, person names, etc. NER in the medical domain is more complicated for two key reasons [24]. First, because of non-standard abbreviations, acronyms, terminology, and several iterations of the same entities, several entities seldom or sometimes fail to appear in the training dataset. Second, text in clinical notes is noisy due to shorter and incomplete sentences and grammatical and typographical errors. Additionally, clinical NER for multiple facilities is more difficult due to potential inter-facility variation compared with single facility data. On the one side, clinical reports involve a paragraph border, incomplete sentences, rendering it impossible to identify complex characters, and sentence boundary detection [25].

A dependency of NLP is the sentence boundary detection (SBD) task is to identify the sentence segments within a text[19]. Good segmentation of sentences will not only improve translation quality but also reduce latency in colonoscopy reports. Most of the existing such as SBD systems SATZ - An Adaptive Sentence Segmentation System[20], proposed by Palmer and Hearst (1997), Disambiguation of Punctuation Marks (DPM), and Automated Speech Recognition (ASR) system are highly accurate on standard and high-quality text, but extracting sentences or detecting the boundary of sentences from a clinical text in colonoscopy reports is a challenging task because clinical language is characterized, in addition to peculiarities like misspellings, punctuation errors and incomplete sentences, by an abundance of acronyms and abbreviations[21].

Neural embedding methods such as word2vec[29] and GloVE[30] have been widely explored over the past five years to construct low-dimensional vector representations of words and passages of text and entity boundary extraction. Neural embeddings are a common group of low-dimensional vector representation methods, phrases, or text (50-500 dimensions) whose values are assigned by a neural network functioning as a black box. However, it isn't easy to view these dimensions in a meaningful way in clinical notes. It takes extensive training and tuning of several parameters and hyperparameters to construct neural embeddings.

This paper describes and evaluates a method for identifying clinical entities and entity boundary extraction from the CRC clinical report. Given a sequence of words, our model represents each word, phrase, or text using a concatenated low-dimensional vector of its corresponding word and character sequence embedding. A near-comprehensive vector representation was built of words and selected bigrams, trigrams, and abbreviations. The word vector and dictionary features are then projected into dense representations of the vectors. These are then fed into the Bidirectional Long Short-Term Memory (Bi-LSTM) to capture contextual features. Lastly, a Conditional Random Field (CRF) is used to capture dependencies between adjacent tags. With the ability of Bio-Bi-LSTM-CRF, we capture detailed colonoscopy-related information within colonoscopy records like indicators, the number of polyps, the size of polyps, and the polyp's location, and the procedure and polyp histology like adenoma and tubulovillous and adenocarcinoma.

Section Methods explains the proposed Bio-Bi-LSTM-CRF in depth. Section Model Evaluation explains about strategies for model evaluation Section Results presents the experiments and results. Section Discussion discusses several important issues of the proposed model. Section Strengths and Limitations explains about strengths and limitations of the model and finally, Section conclusion concludes the current paper and points out potential future work.

## Methods

### Dataset and Annotation

A total of 44,405 de-identified colonoscopy reports, which represent all colonoscopies between 2002 and 2012 on 36,537 subjects, were extracted from four Veterans Affair Medical Centers (Figure 1). These colonoscopy reports contain sections such as indications and findings that contain text input by a colonoscopist. The exact sections can vary between facilities. Only sections that contain text with information on the indication of polyps found during the colonoscopy were retained for analysis. Pathology reports include sections such as impressions and findings with text input by a pathologist. Only sections that contain text with information on polyp pathology were retained for analysis.

Manual annotation of colonoscopy reports provided a gold standard to benchmark the automated methods. A total of 4,000 colonoscopy reports were randomly selected (1,000 from each facility) to annotate sentences. We used 80 percent of the reports as training data, 10 percent as data validation set for hyperparameter tuning, and the remaining 10 percent as test data.

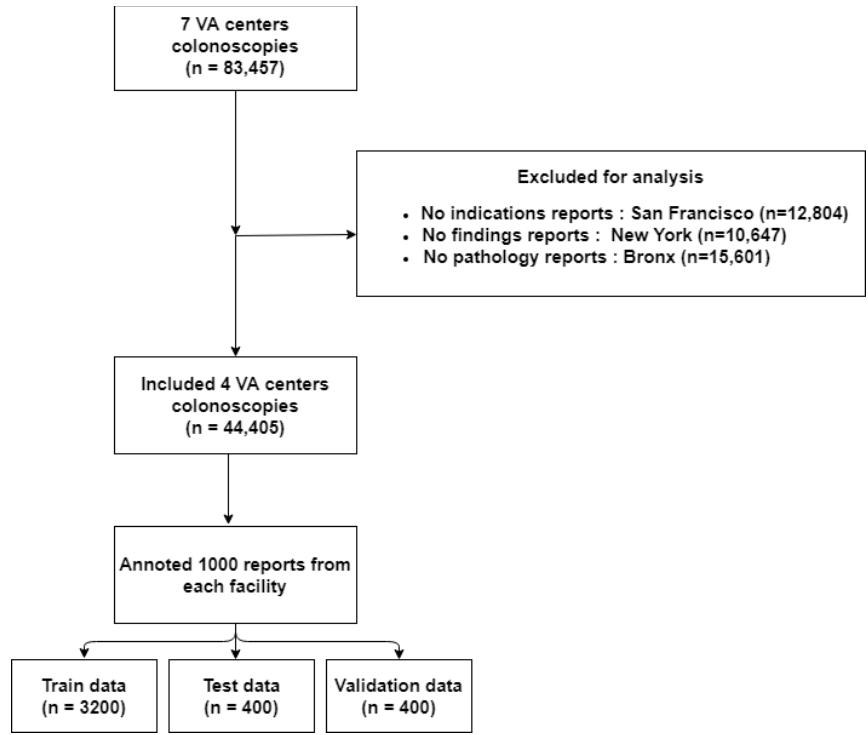

Figure 1. Use of colonoscopy reports for training, validating, and testing automated text extraction methods.

The annotation processes are described in detail[31]. Briefly, two annotators manually marked the sentences devised a list of lexical units (trigger phrases) and corresponding elements (attributes) for each sentence. Based on the existing literature, three common phenotyping tasks were selected. The final list is shown in Table 1.

Table 1. List of the variables of interest from colonoscopy reports

| Lexical Units | Description |
|---|---|
| | |

| Indicators | Reason or indications for colonoscopy examination: Abnormal Computed Tomography; Anemia; Diarrhea; Dyspepsia; Endoscopy; Foreign body removal; Hematochezia; History of colon cancer; Melena; Modification of bowel habits; Neoplasia; Personal history of colonic polyps; polypectomy site; Presence of fecal occult blood; Screening; Screening and surveillance for colon neoplasia; Sigmoidoscopy; Surveillance of patients with neoplastic polyps; Treatment of bleeding from such lesions as vascular; malformation; Ulceration; and Weight Loss |
|---|---|
| Polyp Descriptions | The diameter of the polyp: Small (1-4mm), Medium(5-9mm), or Large (>10mm) <br> Morphology of polyp: sessile, pedunculated, mass <br> The number of polyps: either a numerical value or qualitative description such as many, several, numerous, etc. |
| Location | Proximal: Ileocecal valve, Cecum, Hepatic flexure, Ascending, Transverse, and Splenic flexure <br> Distal location: Descending, Sigmoid, Rectosigmoid, and Rectum |
| Polyp histology | Adenoma, Tubulovillous, Villous, High-grade dysplasia, Adenocarcinoma |

The main components of the entity extraction pipeline are depicted in Figure 2. The pipeline includes (1) data cleaning and pre-processing, (2) sentence boundary detection, (3) vector representation, and (4) the entity extraction model. We used python 3.7 and the open-source packages NLKT, sci-kit–learn, and Tensor Flow that provide various NLP methods, machine learning methods, and neural network modules as described in depth in the sections below.

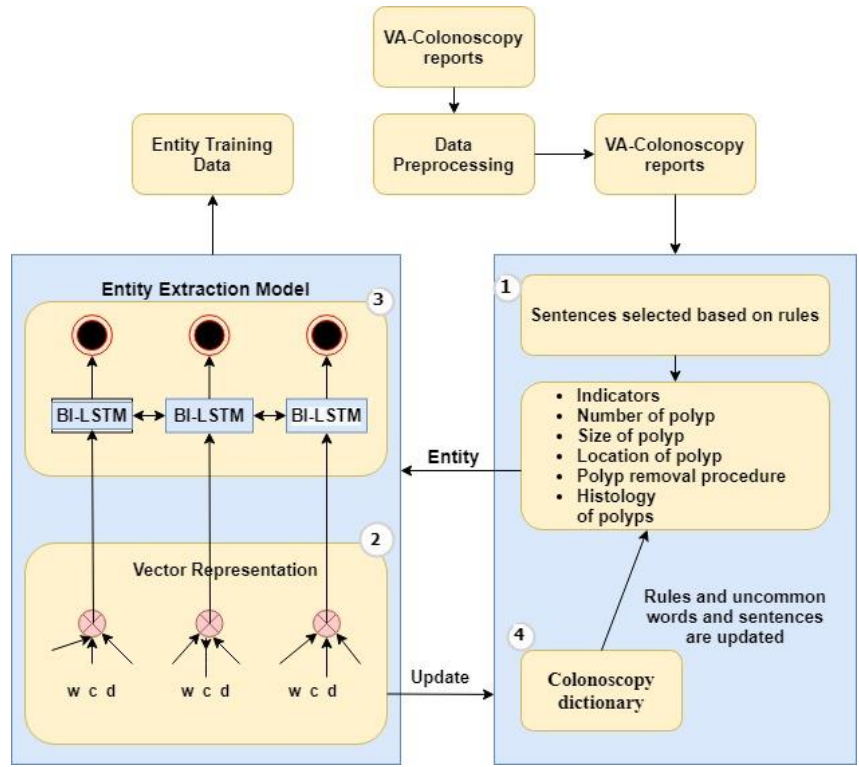

Figure 2. Model architecture. There are four parts to our model: (1) sentence boundary detection, (2) vector representation, (3) entity extraction model, and (4) rule update.

**Data cleaning and Preprocessing**

We pre-processed the text reports to remove all special characters (e.g., commas in a list). Misspellings in our data set included typographical errors (e.g., "cacel" instead of "cecal"), phonetic errors that could be associated with lack of familiarity with medical terms (e.g., byopsi and methastasis), and everyday language errors (e.g., hooooooootsnare). The correction of spelling errors from clinical entries using the MetaMap and auto spell checker by Google's query suggestion service.

**Sentence Boundary Detection**

Sentence segmentation is not a simple task in clinical data. Automatically segmenting clinical free text is an essential first step to information extraction. There are many sentence boundary detection (SBD) packages that mainly relied on punctuation marks and wide spaces. In the most text, sentence boundaries are indicated by a period, but colonoscopy reports frequently contain inter-sentence periods (e.g., "Dr. Patil" or "size of 3.5 cm" or "No. of polyps:") contrasting with typical biomedical text that usually had concepts from standard terminologies. Theoretically, boundary failures can result from standard medical terminologies. To solve this problem, we applied a machine learning approach to identify the sentence boundaries.

To perform sentence boundary detection, it is essential to represent the input sentences with suitable tags. S.F. (S=beginning of a sentence, F= ending of the sentence) format is used tagging in sentence boundary detection. We used the features and keywords listed in Table 4 to identify and tag the sentence boundaries. We used a decision tree for sentence boundary detection classification. This approach is constructed by applying the C4.5 algorithm. The example is shown in Textbox 1.

Textbox 1. Example of clinical note.

> [S]Six sessile polyps were found in the sigmoid colon and in the proximal descending colon[F]. [S]The polyps were 4 to 15 mm in size. All polyps were completely removed by cold snare polypectomy [F].[S] Resection and retrieval were complete[F].[S]Multiple small and large-mouthed diverticulitis were found in the sigmoid colon[F].

The clinical note extracted sentences containing essential lexical units. To avoid duplicate sentences and improve sentence diversity, the sentences were sorted by TF-IDF cosine distance. The sentences were manually de-identified and checked by automatic sentence boundary detection.

**Vector Representation**

Its complexity characterizes information on clinical notes in vocabulary and morphological richness. Therefore, using the recognition of a named entity to capture morphological information from complex medical terms can avoid vocabulary problems and compensate for traditional embedding words. The name entity recognition of clinical colonoscopy notes is usually regarded as a sequence labeling task. Given a sequence of words in a clinical notes sentence, we label each word in the sentence with a BIEOS (Begin, Inside, End, Outside, Single) tag scheme.

For example, in Textbox 2 below, "The mass was circumferential" and "Estimated blood loss was minimal "sentences are not significant. The widely used annotation modes for named entity recognition training data are BIO, BIEO, and BIESO, where B is the beginning of an entity, I reflect the center of an entity, E is the end of an entity, S is a single entity, and O is not an entity.

Textbox 2. Example output after sequence labeling task.

> [S]A sessile polyp was found in the descending colon [F]. [S] The polyp was 8 mm in size [F]. [S] The polyp was removed with a cold snare [F]. [S] Resection and retrieval were complete [F] .[S] [O] Estimated blood loss was minimal [O] [F] . [S] A fungating partially obstructing mass that could be easily traversed was found in the sigmoid colon [F] .[S] [O] The mass was circumferential [O] [F] .[S] The mass measured five cm in length from 25 to 30 cm [F] .

In this study, we create a new vector representation by integrating dictionary function vector due to the uncertainty in the boundary of medical vocabulary terms and the complexity of representing uncommon and unknown entities. Each word's vector representation $e_i$ consists of the word vector $w_i$ and the dictionary feature vector $d_i$ as $e_i = w_i + d_i$ Where + is the concatenation operator, as shown by the diagram, the word vector is constructed as a combination of word embedding and character embedding. To represent character embedding, we use bidirectional LSTM to extract character features and implement certain character features.

**Dictionary feature to vector**

We use the n-gram models to divide the original sentence into text segments based on the word's context. After that, combined with the domain dictionary (colonoscopy) C, we produce a binary value dependent on whether or not the text segments are in C. Dictionary feature vectors are built in various dimensions, based on the number of entity categories in the dictionary. Bi-LSTM converts the final dictionary function vector from the dictionary element.

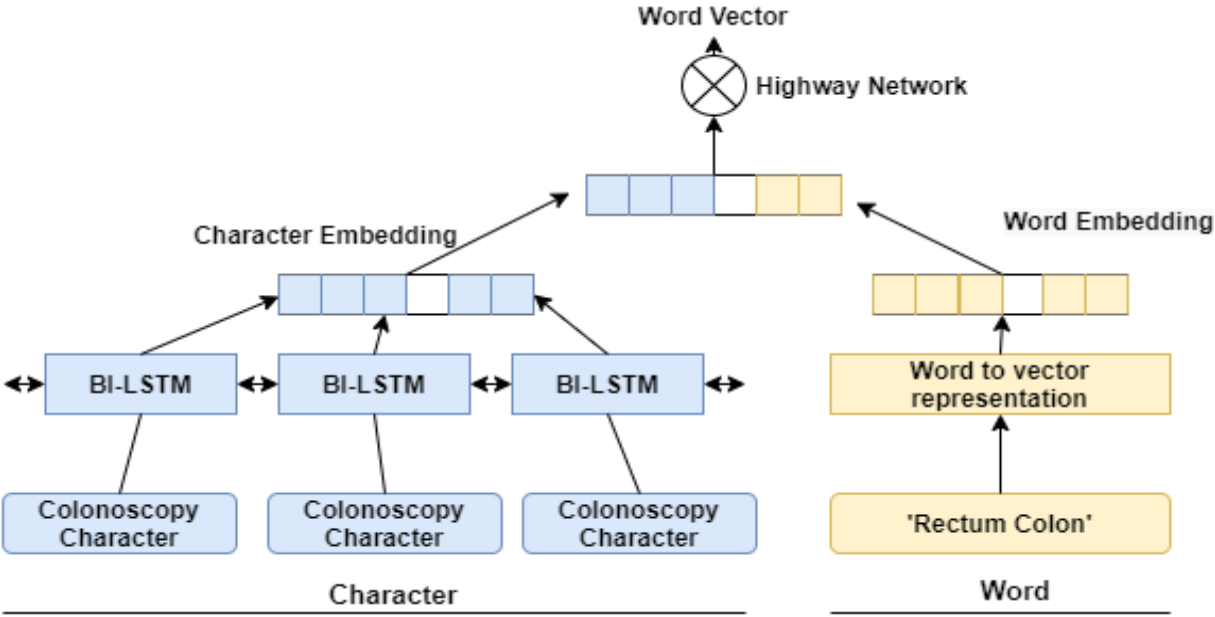

Figure 3. Word vector as a combination of character embedding and word embedding. Each word trains a word vector to the model.

**Entity Extraction Model**

The entity extraction utilizing Bi-LSTM-CRF[32] used word embedding as inputs; we employ embedding character and word embedding. It allows the model to get trained with more feature information. The combined character words and dictionary feature vectors are feed into the model. This step has the following advantages:(1) more details about the features; and (2) better identification of the entity boundary problem. After that, we feed this into a Bi-LSTM model. Instead of splitting the two feature vectors as inputs, three separate Bi-LSTM versions of the proposed model are seen in Figure 3. To obtain a vector of the final character representation, we implemented a Bi-LSTM neural network and added the forward and backward sequences' output vectors. The key aim of utilizing Bi- LSTM is that it functions best to take on long-term dependencies in a phrase, and the bidirectional sequential design provides further advantages by remembering both a word's previous and potential meaning.

CRF Layer: The colonoscopy reports for a sentence, the measurement of the from BI-LSTM neural network obtains a word representation, ignoring neighboring marker dependencies. For example, the tag "I" (Inside) cannot be followed with the tag "B" (Begin) or "E" (End). We use conditional random field (CRF) to determine labels from a sequence of background representations, rather than using it independently for tagging decisions can produce greater tagging accuracy in general.

**Rules for Specific Task**

As we are dealing with unstructured colonoscopy clinical notes from multiple facilities, we use manual rules to fix this problem. The aim of defining rules is to identify sentences that contain certain entities from the text. These rules will be dynamically updated for each facility. We explained the main four rules in Table 2.

Table 2. Description of the six major rules

| Task | Rules |
| --- | --- |
| | |
| Sentence boundary detection | Keywords (e.g., Number of polyps, jar, etc.), period, semicolon, numPreBlanklines, and numPostBlanklines |
| Polyps size | Keywords (e.g., small, medium, large, etc.), number preceeding "CM", or number preceeding "MM" |
| Location of polyps | Keywords (e.g., rectal, sigmoid, etc.) or distance from rectum given in "CM". |
| Word processing | Abbreviations (e.g., "SIG" for sigmoid, "REC" for rectal, etc.) and special word processing (e.g., "Recto-sigmoid" is a combination of "rectal" and "sigmoid", "Tubulovillous" is a combination of "Tubular" and "Villous", etc.) |
| Number of polyps | Keywords (polyps, number, numerics, and sessile) |

**Dataset and Experimental Settings**

As previously mentioned, there are 4,000 manually annotated colonoscopy reports and five kinds of medical named entities of interest: a number of polyps, size of the polyp, location of the polyp, the procedure of removal of polyps, and status of removal as shown in Table 3. There are several sentences to every instance. These sentences are further separated into clauses by using "S" as the beginning of the sentence and "F" as mentioned in section 4.3. We calculate micro-averaged precision, recall, and F1-score for exact matching words and classify sentences accordingly to conceptual meaning from colonoscopy reports. We compare our model Bi-LSTM-CRF with improved NER and sentence boundary detection with the Bi-LSTM-CRF.

Table 3. Statistics of entity and sentence.

| Type | Number of Entities | Number of Sentences |
| --- | --- | --- |
| | | |
| Indication | 4,215 | - |
| Number of polyps | 9,542 | 4,687 |

| Location of polyps | 12,647 | 5,321 |
|---|---|---|
| Size of polyps | 8,654 | 3,549 |
| Polyp removal procedure | 3,657 | 2,648 |
| Type of colorectal cancer | 6,548 | 4,598 |

We used Tensor Flow to complete the CRF module. We first run a bidirectional LSTM-CRF model forward pass for each batch, including the forward pass for both the forward and backward states of LSTM. We divide the learning rate by five if the validation accuracy decreases and used mini-batches of size 100, and training is stopped when the learning rate goes under the threshold of $10^{-8}$. For classifiers, we set hidden layers size to 300; because neural network architectures are usually over-parameterized and vulnerable to overfitting to solve it, we used regularization strategies that have been implemented: Elastic Net regularization has been added to each LSTM hidden layer. The parameters predominantly include tag indices and batch size, where tag indices indicate the number of actual tags, and the batch size reflects the number of batch samples. The parameter settings for the proposed model are shown in Table 4.

Table 4. Parameter setting of the proposed method.

| Parameter | Value |
|---|---|
|  |  |
| Word vector embedding size | 200 |
| Dictionary feature vector embedding size | 100 |
| # hidden neurons for each hidden layer | 300 |
| Batch size | 100 |
| Tag Indices | 4 |
| Learning Rate | 0.005 |
| Number of epochs | 10 |
| Optimizer | Adam optimizer |

## Model evaluation

The proposed Bio-Bi-LSTM-CRF method and the state-of-the-art model Bi-LSTM-CRF were trained on the training subset, and the validation subset was used for tuning the hyper-parameters. Ultimately, the trained models were tested on the test subset, and the following outcomes were reported as the models' performance. We used precision, recall, accuracy, and F1 score to evaluate the Bio-Bi-LSTM-CRF model and Bi-LSTM-CRF model's performance compared with the overall accuracy.

Evaluations of precision, recall, accuracy, and F1 score were performed for each report. For example, all of the polyp numbers, sizes, and locations in a findings report must be correct for it to be considered accurate regardless of the number of polyps. Also, the precision, recall, accuracy, and F1 score were performed at the entity-level (e.g., each polyp size, number, and location) to find the models' performance for each facility. An exact match with the gold standard of manually annotated records is required for each entity versus the record-level where each record is considered separately.

## Results

The proposed Bio-Bi-LSTM-CRF outperformed Bi-LSTM-CRF and achieved higher accuracy by 5.6%, 6.6%, and 19.1% percentage points in the indication, findings, and pathology reports, respectively, as shown in Table 5. The Bio-Bi-LSTM-CRF model's absolute accuracy was relatively high for all three reports: 88.0% for findings reports, 93.5% for indication reports, and 96.5% for pathology reports. The proposed Bio-Bi-LSTM-CRF thus reports a 15.3% increase in precision, a 14.8% increase in recall, and a 10.3% increase in F1 over the state-of-the-art model Bi-LSTM-CRF. The hierarchical influence structure of Bio-Bi-LSTM-CRF aided in reducing noisy text from clinical colonoscopy notes.

Table 5. Comparative results by report between Bi-LSTM-CRF and Bio-Bi-LSTM-CRF model

| Reports | Method | Accuracy | Recall | Precision | F1 |
|---|---|---|---|---|---|
| Indication Reports | Bi-LSTM-CRF | 88.5% (354/400) | 82.5% | 82.5% | 83.2% |
| | Bio-Bi-LSTM-CRF | 93.5% (374/400) | 93.0% | 93.0% | 93.4% |
| Findings Report | Bi-LSTM-CRF | 82.5% (300/400) | 83.5% | 84.0% | 82.7% |
| | Bio-Bi-LSTM-CRF | 88.0% (352/400) | 88.7% | 89.0% | 88.1% |
| Pathology Report | Bi-LSTM-CRF | 81.0% (324/400) | 77.6% | 75.0% | 79.7% |
| | Bio-Bi-LSTM-CRF | 96.5% (386/400) | 96.0% | 96.0% | 96.4% |

Both the Bi-LSTM-CRF and Bio-Bi-LSTM-CRF models performed better in entity identification indications reports, and pathology reports achieved an average of 84.75% and 95% accuracy between facilities, as shown in Table 6. However, they struggled in organizing the findings reports that mentioned characteristics of number polyps and locations of polyp in body and procedure of removal of a polyp, e.g., two diminutive mass polyp, and reports where the status of the location of the polyp was defined, e.g., A single small 3-5mm polyp away from rectum and sigmoid colon' by long-range dependencies and words. In indication reports and pathology reports, there are no long dependency sentences and multiple sentences. Whereas findings have multiples sentences representing the information of polyps. To understand polyps' characteristics, the model needed to look further back in the sentence and relate " polyp" or "sessile."

Table 6. Comparative results by facility between Bi-LSTM-CRF and Bio-Bi-LSTM-CRF model

| Facility | Methods | Accuracy | Recall | Precision | F1 |
|---|---|---|---|---|---|
| | | | | | |
| Albany | Bi-LSTM-CRF<br>Bio-Bi-LSTM-CRF | 83.0% (332/400)<br>92.0% (368/400) | 79.0%<br>91.5% | 81.0%<br>91.0% | 78.0%<br>90.0% |
| Ann Arbor | Bi-LSTM-CRF<br>Bio-Bi-LSTM-CRF | 86.0% (344/400)<br>94.0% (376/400) | 84.0%<br>92.0% | 82.0%<br>93.0% | 85.0%<br>93.5% |
| Detroit | Bi-LSTM-CRF<br>Bio-Bi-LSTM-CRF | 83.0% (332/400)<br>91.0% (364/400) | 83.0%<br>91.0% | 78.0%<br>90.5% | 81.0%<br>91.0% |
| Indianapolis | Bi-LSTM-CRF<br>Bio-Bi-LSTM-CRF | 80.0% (320/400)<br>93.0% (372/400) | 81.0%<br>95.5% | 82.0%<br>95.5% | 84.0%<br>95.5% |

On the other hand, Bio-Bi-LSTM-CRF models entity identification on findings reports correctly. Bi-LSTM-CRF models achieved an average of 92.6% recall, as shown in Table 8. The Bio-Bi-LSTM-CRF model only made a few misclassifications of entities where the polyp location was represented in the form of distance from the rectum (130 cm). The number of polyps was described in an atypical fashion, including uncommon terms such as "semi-mass of large 10mm polyp," which only appeared in one and a few notes respectively in the dataset.

The Bio-Bi-LSTM-CRF model's overall accuracy was between 8.0% and 13.0% points higher than the Bi-LSTM-CRF method in entity identification. The accuracy of the Bio-Bi-LSTM-CRF methods between facilities ranged from 92.0% to 94.0%, as shown in Table 7. The F1-scores of Albany facility, Ann arbor facility, Detroit facility, and Indianapolis facility are 93.7%, 96.3%, and 87.1%; as shown in Table 8 The Bio-Bi-LSTM model achieved the highest overall results for all three facilities. The proposed Bio-Bi-LSTM-CRF thus reports a 12.8% increase in the F1 score. Likely, our Bio-Bi-LSTM-CRF model adapted well to other physicians' reporting style and language to achieve comparable performance in other facilities. For the polyp removal procedure, the performance of the two models of about the same level (at around 90.0%).

In comparison, the Bio-Bi-LSTM-CRF model shows an improvement of 20.9% F1 score over the Bi-LSTM-CRF in a type of CRC information from pathology reports. Bi-LSTM-CRF is struggling to extract entities from long dependences terms and multiple sentences. The weakest performance area was for location information of polyp. There is a large variety of location representation in the dataset. These are also relatively specific and unlikely to occur in the other facility data.

Table 7. Comparative results by performance evaluation between Bi-LSTM-CRF and Bio-Bi-LSTM-CRF model

| Category | Methods | Accuracy | Recall | Precision | F1 |
|---|---|---|---|---|---|
| | | | | | |
| Indicator | Bi-LSTM-CRF<br>Bio-Bi-LSTM-CRF | 83.5% (334/400)<br>93.5% (374/400) | 82.5%<br>93.1% | 82.5%<br>93.0% | 83.2%<br>93.4% |
| Number of polyps | Bi-LSTM-CRF<br>Bio-Bi-LSTM-CRF | 80.0% (320/400)<br>85.0% (340/400) | 76.0%<br>88.0% | 86.0%<br>89.0% | 77.0%<br>89.0% |
| Size | Bi-LSTM-CRF<br>Bio-Bi-LSTM-CRF | 78.0% (312/400)<br>81.0% (324/400) | 78.0%<br>88.0% | 80.5%<br>89.0% | 79.0%<br>90.0% |
| Location of polyp | Bi-LSTM-CRF<br>Bio-Bi-LSTM-CRF | 82.0% (328/400)<br>94.0% (376/400) | 83.0%<br>86.0% | 78.0%<br>87.0% | 82.0%<br>85.0% |
| Polyp removal procedure | Bi-LSTM-CRF<br>Bio-Bi-LSTM-CRF | 89.0% (356/400)<br>92.0% (368/400) | 94.0%<br>94.0% | 93.0%<br>93.0% | 84.0%<br>92.0% |
| Type of CRC | Bi-LSTM-CRF<br>Bio-Bi-LSTM-CRF | 81.0% (324/400)<br>96.5% (386/400) | 77.6%<br>96.0% | 75.0%<br>96.0% | 79.7%<br>96.4% |

## Discussion

This study proposed a Bio-Bi-LSTM-CRF model for efficient and accurate entity extraction from colonoscopy report notes and free-text medical narratives. The novel Bio-Bi-LSTM-CRF accurately analyzed a large sample of colonoscopy reports. Our findings demonstrate a clear improvement in accuracy, precision, and recall within a highly regarded academic healthcare system. Bio-LSTM-CRF model consistently obtained better entity extraction accuracy than a Bi-LSTM-CRF model with identical feature sets. Over experiments, it was found that the addition of sentence boundary detection, rules, and domain dictionary in the tasks could significantly boost the accuracy, thus proving the efficacy of merging practices and domain dictionary in entity extraction tasks.

Our results highlight several advantages of using Bio-Bi-LSTM-CRF in routine quality measurement using data in EHRs. The critical advantage of Bio-Bi-LSTM-CRF is that it is economically feasible. It would be expensive and time-consuming to review tens of thousands of colonoscopies reports manually. Another advantage is that Bio-Bi-LSTM-CRF allows providers to continue to use natural narrative when describing patient care. There has been criticism that structured note systems in current EHRs force providers to create unnatural and overly structured notes, which take extra time to develop and impede communication because these notes are difficult to read [33].

Compared to the direct and implicit term metrics, the neural embeddings were observed to have the inferior performance on the biomedical term similarity/relatedness benchmarks, as assessed both by our tests (against several different implementations of word2vec that other groups have developed for similar tasks) and those reported by others (see Results). However, we acknowledge that that word2vec performance can be optimized for specific tasks by adjusting different choices of parameters and hyperparameters, different neural network architectures, and other ways of handling multi-word phrases [34, 35], and we did not test all possible implementations. Thus, we do not claim that our novel metrics necessarily perform better than neural embeddings in general. We decided to include only the most essential, informative words and phrases in our vocabulary in contrast to the word2vec based embeddings that encompass a much larger vocabulary. We feel that the relatively prevalent words and phrases are likely to be the most valuable for representing biomedical articles. However, it is worth exploring the effects of different vocabulary restrictions, especially when applying our vectors to other corpora, domains, and tasks.

**Strengths and Limitations**

In many aspects, our research is novel. First, to our knowledge, this is only the research to examine the validity of NLP for multiple center data to investigate indications, findings, and pathology laboratory results of dictated consultation notes for this purpose [36, 37]. Secondly, NLP research in CRC clinical notes concentrates exclusively on a single clinical condition, such as tumor presence [26]. Our analysis is significantly more comprehensive than other more widely reported studies, looking at four different medical centers. There are multiple drawbacks to our research. In our initial training dataset (n=4,000), our choice to randomly sample resulted in CRC cases' under-sampling. As a result, the model was never trained on this feature and subsequently performed poorly during the final test set for this feature, possibly underestimating a properly trained model's effectiveness. This suggests that a real-world application of this technology may require a more purposive sampling strategy than our random sampling approach. Although the inclusion of many bigrams, trigrams, and abbreviations should reduce the problem of term ambiguity to some extent, the vector representations of words represent an overall average across word instances and different word senses. The implicit term metrics have relatively few parameters, which we have attempted to set at near-optimal values. However, the choice of 300-dimensions for the vector representations, and the particular weighting schemes for the similarity of vectors, may be regarded as best guesses and might be tuned further for optimal performance in specific tasks. However, this study's goal was to establish the feasibility of using a deep learning model to extract clinical features from dictated consult notes and inform the approach to more extensive future studies.

**Conclusion**

Our proposed Bio-BI-LSTM-CRF improved the perception of unknown entities, which means that the model should have a better capability to deal with emerging medical concepts and multicenter data without extra training resources. The pre-processing of the word-level input of the model with external word embeddings allowable to improve performance further and achieve state-of-the-art for the colonoscopy clinical notes entity extraction. This idea could not only be applied in medical entity extraction but also other medical named entity recognition applications such as drug names and adverse drug reactions, as well as named entity recognition tasks in different fields.

# References

1. Alblawi, A.S. and A.A. Alhamed. Big data and learning analytics in higher education: Demystifying variety, acquisition, storage, NLP and analytics. in 2017 IEEE Conference on Big Data and Analytics (ICBDA). 2017. IEEE.
2. Denecke, K. Extracting medical concepts from medical social media with clinical NLP tools: a qualitative study. in Proceedings of the fourth workshop on building and evaluation resources for health and biomedical text processing. 2014. Citeseer.
3. Strubell, E., A. Ganesh, and A. McCallum, Energy and policy considerations for deep learning in NLP. arXiv preprint arXiv:1906.02243, 2019.
4. Jiang, F., et al., Artificial intelligence in healthcare: past, present and future. Stroke and vascular neurology, 2017. **2**(4).
5. Dinov, I.D., Methodological challenges and analytic opportunities for modeling and interpreting Big Healthcare Data. Gigascience, 2016. **5**(1): p. s13742-016-0117-6.
6. Mishuris, R.G. and J.A. Linder, Electronic health records and the increasing complexity of medical practice: "it never gets easier, you just go faster". Journal of general internal medicine, 2013. **28**(4): p. 490-492.
7. Polnaszek, B., et al., Overcoming the Challenges of Unstructured Data in Multi-site, Electronic Medical Record-based Abstraction. Medical care, 2016. **54**(10): p. e65.
8. Meystre, S., Clinical Data Reuse or Secondary Use: Current Status and Potential Future Progress (2017).
9. Jemal, A., et al., Cancer statistics, 2010. CA: a cancer journal for clinicians, 2010. **60**(5): p. 277-300.
10. Center, M.M., A. Jemal, and E. Ward, International trends in colorectal cancer incidence rates. Cancer Epidemiology and Prevention Biomarkers, 2009. **18**(6): p. 1688-1694.
11. Levin, B., et al., Screening and surveillance for the early detection of colorectal cancer and adenomatous polyps, 2008: a joint guideline from the American Cancer Society, the US Multi-Society Task Force on Colorectal Cancer, and the American College of Radiology. Gastroenterology, 2008. **134**(5): p. 1570-1595.
12. Phillips, K.A., et al., Trends in colonoscopy for colorectal cancer screening. Medical care, 2007: p. 160-167.
13. Winawer, S.J., et al., prevention of colorectal cancer by colonoscopic polypectomy. New England Journal of Medicine, 1993. **329**(27): p. 1977-1981.
14. Denny, J.C., et al., Extracting timing and status descriptors for colonoscopy testing from electronic medical records. Journal of the American Medical Informatics Association, 2010. **17**(4): p. 383-388.
15. Goldberg, Y., Neural network methods for natural language processing. Synthesis lectures on human language technologies, 2017. **10**(1): p. 1-309.
16. Jagannatha, A.N. and H. Yu. Structured prediction models for RNN based sequence labeling in clinical text. in Proceedings of the conference on empirical methods in natural language processing. conference on empirical methods in natural language processing. 2016. NIH Public Access.
17. Pons, E., et al., Natural language processing in radiology: a systematic review. Radiology,

processing. American Journal of Gastroenterology, 2015. **110**(4): p. 543-552.
37. Lee, J.K., et al., Accurate identification of colonoscopy quality and polyp findings using natural language processing. Journal of clinical gastroenterology, 2019. **53**(1): p. e25.
11